\documentclass{article}
\usepackage{
    spconf,
    amsmath,
    graphicx,
    multirow,
    multicol, 
    arydshln, 
    hyperref,
    svg,
    cite,
    float,
    caption,
    enumitem,
    fancyhdr 
}

\hypersetup{ 
    colorlinks=true, 
    linkcolor=black, 
    citecolor=black, 
    urlcolor=blue,
    linkbordercolor={1 1 1},
    pdfborder={0 0 0},
}

\title{ContextLoss: Context information for topology-preserving segmentation}
\name{Benedict Schacht$^{1,2}$, Imke Greving$^{2}$, Simone Frintrop$^{1}$, Berit Zeller-Plumhoff$^{2,3}$, Christian Wilms$^{1}$}
\address{$^1$University of Hamburg, $^2$Helmholtz-Zentrum Hereon, $^3$University of Rostock}

\fancypagestyle{FirstPage}{
\lfoot{\small \copyright 2025 IEEE. Personal use of this material is permitted. Permission from IEEE must be obtained for all other uses, in any current or future media, including reprinting/republishing this material for advertising or promotional purposes, creating new collective works, for resale or redistribution to servers or lists, or reuse of any copyrighted component of this work in other works.} 
}

\begin{document}
\maketitle
\begin{abstract}
In image segmentation, preserving the topology of segmented structures like vessels, membranes, or roads is crucial. For instance, topological errors on road networks can significantly impact navigation.
Recently proposed solutions are loss functions based on critical pixel masks that consider the whole skeleton of the segmented structures in the critical pixel mask. We propose the novel loss function ContextLoss (CLoss) that improves topological correctness by considering 
topological errors with their whole context in the critical pixel mask.
The additional context improves the network focus on the topological errors. 
Further, we propose two intuitive metrics to verify improved connectivity due to a closing of missed connections.
We benchmark our proposed CLoss on three public datasets~(2D \& 3D) and our own 3D nano-imaging dataset of bone cement lines. 
Training with our proposed CLoss increases performance on topology-aware metrics and repairs up to 44\,\% more missed connections than other state-of-the-art methods.
We make the code publicly available\footnote{\href{https://gitlab.com/Benedict_S/contextloss}{https://gitlab.com/Benedict\_S/ContextLoss}}
\footnote{\href{https://dx.doi.org/10.60864/jkny-a610}{Supplement: https://dx.doi.org/10.60864/jkny-a610}}.
\end{abstract}
\section{Introduction}
\begin{figure}[ht] 
\centering   
\includegraphics[width=0.48\textwidth]{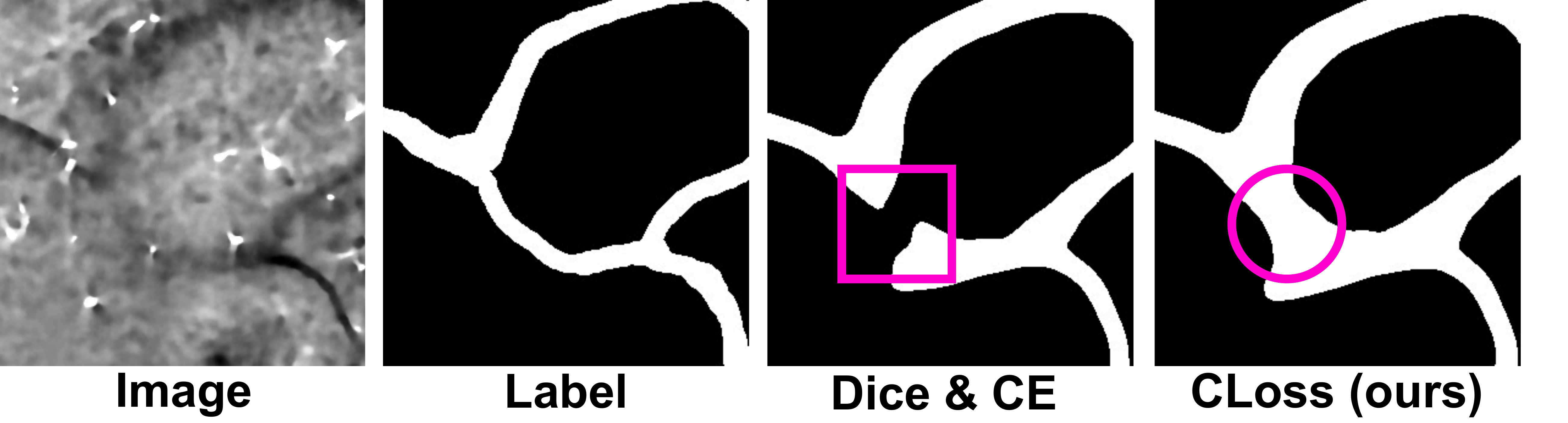}
\caption{\textbf{Motivation.} Image segmentation in bone research, where our method improves topological correctness over training with Dice and Cross-Entropy loss (Dice \& CE).}
\label{fig:Motivation}
\end{figure}

\thispagestyle{FirstPage}
\label{sec:intro}
Loss functions are a core component of neural networks, which determine what the network is optimized for. In image segmentation, 
this translates directly to how well the predictions of the network can resemble the structural properties of the ground truth.
The two most commonly used loss functions in image segmentation are the pixel-wise loss functions Dice loss ($L_{Dice}$) and Cross-Entropy loss ($L_{CE}$).
However, these losses optimize for a maximum overlap of predictions and ground truth by weighting all pixel contributions equally \cite{MA2021102035, Do-we-really-need-dice2024}. 
This is a problem if the topology is of particular interest because the topology is not explicitly considered. This can promote topological errors, eg. in downstream applications related to the segmentation of road systems, blood vessels, or bone features, see~Fig.~\ref{fig:Motivation}. 
\\
\\
Related work \cite{clDice-Shit-2021, Menten2023skeletonization, skeleton-recall2024 ,hu2019topology,warping-NEURIPS2022_98143953, clCE_2024} implements topology sensitivity by applying a pixel-wise loss function ($L_{pixel}$) like $L_{Dice}$ or $L_{CE}$ on a critical pixel mask of topologically important pixels and combining it with an additional $L_{pixel}$ that is applied on all pixels.
Applying $L_{pixel}$ on the critical pixel mask ensures higher network attention on the critical pixels than on the normal pixels.
Related work defines the critical pixel mask by including the complete morphological skeletons of ground truth and predictions \cite{clDice-Shit-2021,Menten2023skeletonization, clCE_2024}, or a slightly dilated skeleton \cite{skeleton-recall2024}. \cite{clCE_2024,skeleton-recall2024,Menten2023skeletonization} extend \cite{clDice-Shit-2021} mainly by different $L_{pixel}$ or removing small structures during training.
\\
\\
While the morphological skeleton is an excellent measure for topological correctness, the complete skeleton typically includes many already correctly predicted pixels. However, the network must be trained to learn a better representation of the topological errors \cite{hu2019topology, warping-NEURIPS2022_98143953}.
\cite{hu2019topology} and \cite{warping-NEURIPS2022_98143953} identify critical pixel masks at the topological errors without skeletonization at the cost of high runtime. However, their identified critical pixels 
are either singular pixels \cite{hu2019topology} or a line of 1-pixel width \cite{warping-NEURIPS2022_98143953} in the center of the topological errors. 
Considering only a few singular pixels in the critical pixel mask is a problem because the network can obtain better results if more pixels are included in the mask \cite{gotkowski2024embarrassinglysimplescribblesupervision}. We argue that \cite{hu2019topology} and \cite{warping-NEURIPS2022_98143953} are missing important context information by including only few critical pixels.
\\
\\
In this paper, we propose the novel loss function CLoss for topology-preserving segmentation. CLoss features an extensive critical pixel mask of the topological errors and particularly considers more context pixels of the topological errors. 
The additional context pixels are acquired by combining skeletonization and the distance transform. 
Utilizing skeletonization in the critical pixel selection makes CLoss naturally fast and compatible with 3D data.
Our loss can be used on 2D and 3D data with any arbitrary segmentation network to improve topological correctness in a plug-and-play manner. 
Additionally, we combine our proposed CLoss with a dedicated training strategy to exploit topological post-processing. For validation, we isolate the contribution of our proposed critical pixel mask from the impact of $L_{pixel}$. We benchmark on four datasets, including our own novel 3D bone Cement Line Dataset (CLD). CLD contains boundary layers of bone tissue which challenge segmentation algorithms with low contrast, diffuse borders, and image artifacts. Further, we propose two intuitive metrics to verify improved connectivity due to a closing of 
missed connections, which complement existing metrics and are more robust to artifacts from evaluation.
In this paper, our contributions can be summarized as follows:
\begin{itemize}[nosep]
\item
We propose the novel topology-preserving loss function CLoss which utilizes a critical pixel mask to consider the whole context of topological errors~(Sec.~\ref{ssec:CLoss}). 
\item
We propose a new 3D bone Cement Line Dataset (CLD) for topology-preserving segmentation~(Sec.~\ref{ssec:Cement line dataset}). 
\item
We propose two metrics which are sensitive to
missed connections~(Sec.~\ref{ssec:Proposed Metrics}).
\end{itemize}
\section{Methodology}
\label{sec:Method}
\subsection{ContextLoss}
\label{ssec:CLoss}
\begin{figure}[ht] 
\centering 
\includegraphics[width=0.48\textwidth]{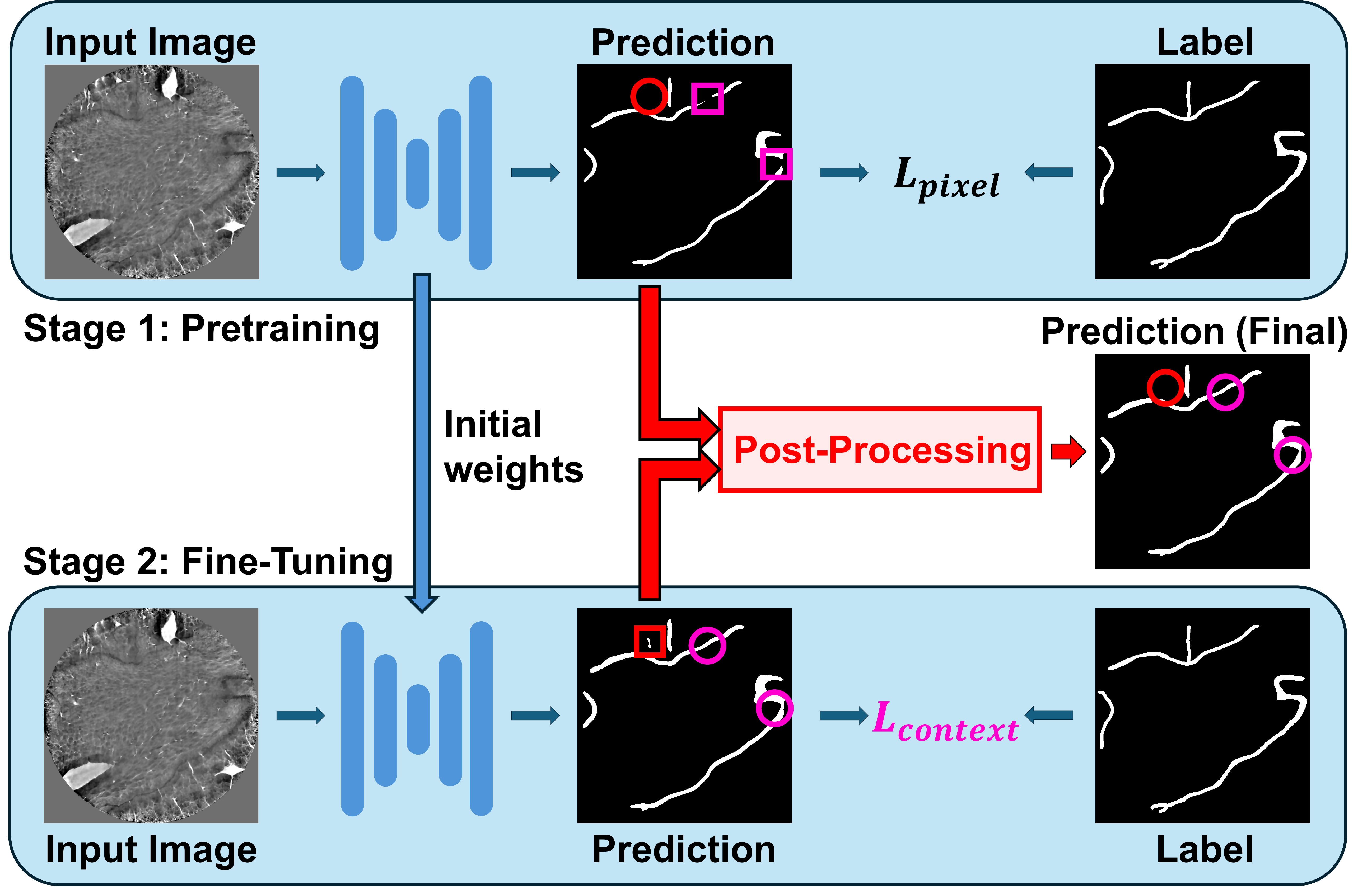} 
\caption{\textbf{Our proposed method.} An arbitrary segmentation network is pretrained with a standard pixel-wise loss ($L_{pixel}$) and then fine-tuned with our topology-sensitive CLoss ($L_{context}$). We combine the inference results of the pretraining for topological post-processing with the inference results from the fine-tuning.  
Our method improves topological correctness (circles) of topological errors (squares).}
\label{fig:CLoss} 
\end{figure}

\begin{figure*}[ht] 
\centering 
\includegraphics[width=\textwidth]{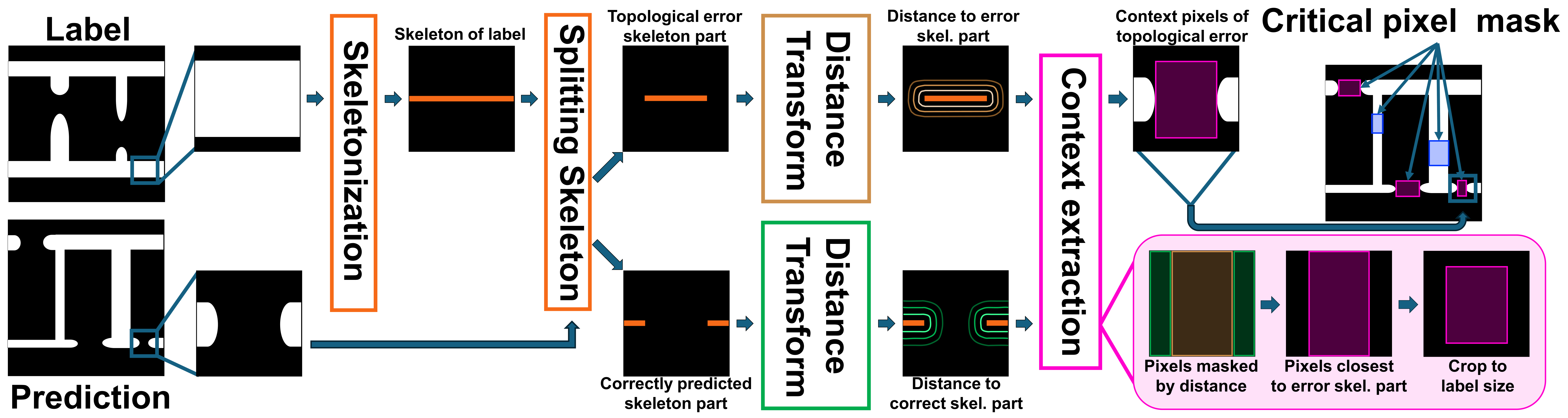} 
\caption{\textbf{Workflow for extracting the critical pixel mask.} CLoss considers all pixels of missed connections (pink mask) and false positive connections (blue mask) as critical pixels, which captures the complete context of the topological errors. We display the critical pixel extraction for the pink mask.
Extracting the blue mask is analogue to the pink mask extraction, just with interchanged label and prediction notation, and cropping to prediction size instead of label size during context extraction.}
\label{fig:crit_pxl_mask} 
\end{figure*}

We propose the novel loss function CLoss to promote topology-preserving segmentation with arbitrary segmentation networks. 
CLoss extracts a critical pixel mask of topologically important locations, which is then considered in the total cost function.
In addition, we introduce a pretraining and topological fine-tuning strategy. The results of both stages are combined in a post-processing to generate the final results, see Fig.~\ref{fig:CLoss}. 
\\
\\
\textbf{Critical pixel mask.} The acquisition of the critical pixel mask is illustrated in Fig.~\ref{fig:crit_pxl_mask}. The critical pixel mask contains the complete context of all topological errors. We determine the skeleton at the
topological critical locations, which is then extended to include all context pixels, using the distance transform. Missed connections are processed as displayed, which we describe below. False positive connections are processed similarly, as mentioned in the caption of~Fig.~\ref{fig:crit_pxl_mask}.
\\
\\
We first obtain the skeleton of the label. Then, we obtain the skeleton part at the topological error and the correctly predicted skeleton part by splitting the label skeleton with the prediction mask. Then, we apply the distance transform to both splitting results, which calculates the distance of all pixels to the respective split skeleton part. Based on the results of the distance transforms, the context extraction keeps only the pixels closest to the skeleton split of the topological error~(see bottom right in Fig.~\ref{fig:crit_pxl_mask}). 
The context pixel selection is further cropped to label size to focus on the context pixels of the topological error. The resulting critical pixel mask ($M$) contains all context pixels of missed and false positive connections.
See our pseudo-code in the supplement for details~(Sec.~\ref{ssec: Appendix, Critical pixel mask}).
\\
\\
\textbf{Cost function.}
Our proposed CLoss is based on a pixel-wise loss function ($L_{pixel}$), which is adapted for topology-sensitivity.
For $L_{pixel}$, we choose the loss combination of $L_{Dice}$ and $L_{CE}$ \begin{equation} 
L_{pixel} = (1-\alpha) L_{Dice} + \alpha L_{CE},
\end{equation}
with $\alpha \in [0,1]$ as a weight. For this work, we use $\alpha=0.5$ for a more robust performance compared to individual $L_{Dice}$ and $L_{CE}$ \cite{MA2021102035,nnU-Isensee-2021}. 
We combine $L_{pixel}$ with the critical pixel mask $M$ to our proposed CLoss
\begin{equation} 
L_{context} = (1-\gamma) L_{pixel} + \gamma  L_{pixel} \odot M ,
\label{eq: Lfinetuning}
\end{equation}
with the Hadamard product $\odot$. $\gamma \in [0,1]$ is a weight to adjust the impact of the topology-sensitive term that contains $M$. 
\\
\\
\textbf{Training with CLoss.}
First, a segmentation network is trained with regular $L_{pixel}$. Subsequently, CLoss is used for fine-tuning to improve topological correctness, see Fig.~\ref{fig:CLoss}. 
Empirically, we found a fine-tuning length with 5\% of the pretraining length to be most efficient. We include ablations in the supplement (Sec.~\ref{ssec: Appendix, additional quantitative results}). 
\\
\\
\textbf{Topological Post-Processing.}
We apply topological post-processing to the fine-tuning result, which only keeps structures already present in the pretraining, see~Fig.~\ref{fig:CLoss}. Intuitively, this ensures topological fine-tuning without adding new foreground structures. We include a formula and more details in the supplement (Sec.~\ref{ssec:micro-noise})

\subsection{Proposed Metrics}
\label{ssec:Proposed Metrics}
Empirically, missed connections (gaps) in the predictions are 
far
more frequent than false positive connections in our used datasets, see Fig.~\ref{fig:Qualitative}. Hence, we propose additional metrics to better evaluate gap closing, adding desirable topological information~\cite{Maier-Hein2024, berger2024pitfallstopologyawareimagesegmentation}.
Our proposed metrics complement existing metrics, which usually evaluate for gaps and false positive connections simultaneously. However, our metrics are more robust against evaluation artifacts, see Fig.~\ref{fig:Metric_artifact}. We include 
a qualitative example 
in the supplement (Sec.~\ref{ssec: Appendix Qualitative example of proposed Metrics}).
\\
\\
\textbf{e$_0$-Gt.} 
Recent work \cite{clDice-Shit-2021,Menten2023skeletonization, hu2019topology,warping-NEURIPS2022_98143953,skeleton-recall2024,lux2024topographefficientgraphbasedframework, Li2023} reports the betti matching error e$_0$~$=| \beta^{X_{bin}}_0 - \beta^{Y}_0|$, 
with the number of connected components of the binary predictions $\beta^{X_{bin}}_0$ and ground truth $\beta^{Y}_0$. e$_0$ is typically evaluated on patches, 
so opposite differences in $\beta_0$ at different locations don't balance each other out.
However, patch evaluation can lead to cropping artifacts with thick predictions, see~Fig.~\ref{fig:Metric_artifact}.
We propose 
\begin{equation} 
\text{e$_0$-Gt}= |\beta_{0}^{X_{bin} \odot Y} - \beta^{Y}_0|,
\end{equation}
which is similar to e$_0$, but the binary predictions are additionally masked with the ground truth. 
Therefore, e$_0$-Gt is not affected by artifacts from thickened predictions. A lower score indicates more closed gaps.
\begin{figure}[ht] 
\centering 
\includegraphics[width=0.48\textwidth]{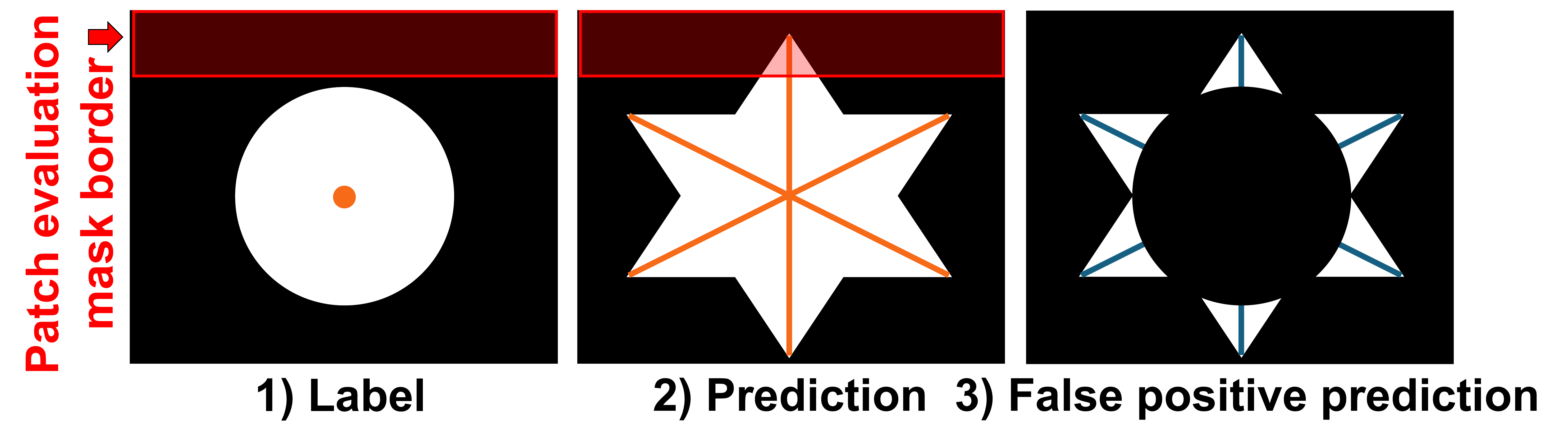}   
\caption{\textbf{Evaluation artifacts.} 1) has the same topology (e$_0$) as 2). 
However, the higher thickness of 2), compared to 1), leads to a cropping artifact in the patch-based evaluation (red area), which falsely suggests an unfavorable e$_0$. Our e$_0$-Gt is robust to this artifact from patch evaluation.
Moreover, the complex contour of 2) evokes many additional skeleton pixels (orange). Paired with the higher thickness of 2), this leads to multiple false positive skeleton pixels (blue), resulting in a bad clDice score. Our AGS is robust to this contour artifact.}
\label{fig:Metric_artifact} 
\end{figure}
\begin{table*}[ht]
    \caption{\textbf{Quantitative results.} Our dataset, proposed method and proposed metrics are indicated in italic. 
    }
    \label{tab:Quantitative results}
    \centering
    \begin{tabular}{ccc|ccccccc}
        \hline
        \hline
        Dataset & Method & Weight $\gamma$ & Dice↑ & clDice↑ \cite{clDice-Shit-2021} & e↓ & e$_1$↓ & e$_0$↓ & \textit{e$_0$-Gt}↓ & \textit{AGS}↑ \\
        \hline
        \hline
         \multirow{8}{*}{\centering Roads} & nnU-Net \cite{nnU-Isensee-2021} &  & 79.69 & 89.34 & 1.181 & 0.895 & 0.286 & 0.702 & 86.46\\
         \cline{2-10}
         &Dice \& CE &  & 79.77 & 89.37 & 1.156 & 0.949 & 0.207 & 0.699 & 86.41\\
         &clDice \cite{clDice-Shit-2021} & 0.5 & 79.50 & 89.11 & 1.126 & 0.897 & 0.230 & 0.710 & 87.07\\
         &Compound clDice & 0.3 & 79.73 & 89.34 & 1.108 & 0.909 & 0.199 & 0.673 & 87.05\\
         \cdashline{2-10}
         &\textit{CLoss} & \textit{0.08} & \textbf{79.82} & \textbf{89.47} & 1.065 & 0.880 & 0.185 & 0.656 & 87.70\\
         && \textit{0.1} & 79.57 & 89.21 & \textbf{0.990} & 0.810 & \textbf{0.180} & 0.617 & 87.88\\
         && \textit{0.2} & 79.12 & 89.13 & 0.994 & \textbf{0.788} & 0.205 & \textbf{0.494} & \textbf{89.48}\\
        \hline
        \hline
         \multirow{8}{*}{\centering HRF-Retina} & nnU-Net \cite{nnU-Isensee-2021} &  & \textbf{82.33} & 81.88 & 0.528 & 0.281 & 0.247 & 2.742 & 77.75\\
         \cline{2-10}
         &Dice \& CE &  & 82.23 & 81.69 & 0.554 & 0.294 & 0.260 & 2.780 & 77.38\\
         &clDice \cite{clDice-Shit-2021} & 0.5 & 82.15 & 83.09 & 0.426 & 0.256 & 0.170 & 2.475 & 82.23\\
         &Compound clDice & 0.5 & \textbf{82.33} & 82.98 & 0.405 & 0.250 & \textbf{0.155} & 2.429 & 81.25\\
         \cdashline{2-10}
         &\textit{CLoss} & \textit{0.08} & 82.22 & 83.20 & 0.447 & 0.259 & 0.187 & 2.347 & 82.28\\
         && \textit{0.1} & 82.20 & 83.27 & 0.449 & 0.262 & 0.188 & 2.469 & 82.61\\
         && \textit{0.2} & 80.57 & \textbf{85.06} & \textbf{0.397} & \textbf{0.225} & 0.172 & \textbf{1.378} & \textbf{90.39}\\
        \hline
        \hline
         \multirow{8}{*}{\centering Vessap} & nnU-Net \cite{nnU-Isensee-2021} &  & 92.89 & \textbf{95.83} & 28.95 & 1.413 & 27.54 & 14.28 & 96.03\\
         \cline{2-10}
         &Dice \& CE&  & \textbf{93.13} & 95.81 & 26.67 & 1.280 & 25.39 & 13.15 & 96.26\\
         &clDice \cite{clDice-Shit-2021} & 0.4 & 92.70 & 94.86 & 29.00 & 1.240 & 27.76 & 9.44 & 97.74\\
         &Compound clDice & 0.1 & 93.11 & 95.80 & 26.10 & \textbf{1.200} & 24.90 & 11.68 & 97.00\\
         \cdashline{2-10}
         &\textit{CLoss} & \textit{0.08} & 92.91 & 95.01 & 26.09 & 1.227 & 24.86 & 9.97 & 97.85\\
         && \textit{0.1} & 92.28 & 94.57 & 27.02 & 1.300 & 25.72 & 9.32 & 98.03\\
         && \textit{0.2} & 91.69 & 93.13 & \textbf{25.07} & 1.827 & \textbf{23.24} & \textbf{6.22} & \textbf{99.04}\\
        \hline
        \hline
         \multirow{8}{*}{\centering \textit{CLD}} & nnU-Net \cite{nnU-Isensee-2021} &  & 69.67 & 83.97 & 3.243 & 1.085 & 2.158 & 2.493 & 79.19\\
         \cline{2-10}
         &Dice \& CE & & 70.27 & 85.23 & 3.422 & 1.075 & 2.347 & 2.447 & 82.02\\
         &clDice \cite{clDice-Shit-2021} & 0.5 & 70.88 & 86.22 & 3.374 & 1.042 & 2.333 & 1.825 & 87.11\\
         &Compound clDice & 0.5 & 70.61 & 85.45 & 3.165 & 1.065 & 2.099 & 2.150 & 82.66\\
         \cdashline{2-10}
         &\textit{CLoss} & \textit{0.08} & \textbf{70.98} & 86.22 & 3.158 & 1.063 & 2.095 & 2.061 & 84.47\\
         && \textit{0.1} & 70.75 & 86.26 & 3.109 & 1.069 & 2.040 & 2.040 & 84.44\\
         && \textit{0.2} & 70.44 & \textbf{86.83} & \textbf{3.020} & 1.047 & \textbf{1.973} & 1.899 & 85.98\\
         &\textit{CLoss (Dice)} & \textit{0.08} &69.43&86.40&3.275&\textbf{0.969}&2.306&\textbf{1.525}&\textbf{90.51}\\
         \hline
         \hline
    \end{tabular}
\end{table*}
\\
\\
\textbf{AGS - Accuracy on Ground Truth Skeleton.} 
Recent work \cite{clDice-Shit-2021,lux2024topographefficientgraphbasedframework,skeleton-recall2024, clCE_2024} reports the clDice metric, which evaluates the intersection of skeletonized prediction ($S_X$) and skeletonized ground truth ($S_Y$). clDice complements e$_0$, as it is sensitive to the partial reparation of gaps and false positive connections, which $\beta_0$-based metrics are not.
However, $S_X$ is overly sensitive to skeletonization artifacts related to a complex contour or surface structure, see Fig.~\ref{fig:Metric_artifact}.
We propose 
\begin{equation} 
\text{AGS}= \frac{ sum\big(X_{bin} \odot S_Y\big)}{sum\big(S_Y \big) },
\end{equation}
which measures the accuracy of the binary predictions ($X_{bin}$) on $S_Y$. AGS is robust to artifacts in $S_X$, since AGS doesn't utilize $S_X$. A higher score indicates a better closing of gaps.
\section{Cement line dataset (CLD)}
\label{ssec:Cement line dataset} 
Cement lines form and separate structural units in the bone, and are thought to play an essential role in crack propagation, energy absorption, and bone mineralization \cite{bsu2024}. Their topology is a key property for their function \cite{bsu2024, Grunewald:ro5034}, therefore automated topology-preserving segmentation creates a substantial benefit in this research field to improve the understanding of cement lines. 3D nano-imaging\footnote{Synchrotron radiation-based full-field transmission X-ray
nanotomography operated in Zernike phase contrast microscopy mode \cite{TXM_2020}.} of cement lines can specifically improve our understanding of cement lines on the nano-scale.
Unfortunately, nano-imaging quality is degraded by noise and various imaging artifacts \cite{CL-SEFA2023154}. Moreover, cement lines have low contrast with respect to their surroundings and diffuse borders. 
Our CLD consists of 17 3D images of shape 1024² $\times$ 600, featuring a voxel size of~(45.6\,nm)³. More details on the annotation process and image acquisition are included in the supplement (Sec.~\ref{ssec: Appendix, Cement line dataset}).
The characteristics of CLD are distinctly different from current 2D topology-benchmark datasets like Roads \cite{dataset-MassRoads-2013} and HRF-Retina \cite{dataset-HRF-2013}, where the foreground class usually has adequate contrast and clear borders. CLD also features a layer/membrane-like foreground structure that is not included in the 3D topology-benchmark dataset Vessap \cite{dataset-Vessap-2019}, which has a tubular structure.

\section{Experiments}
\label{sec:Experiments}
\textbf{Datasets.} 
Following other work on topology-preserving segmentation \cite{clDice-Shit-2021,Menten2023skeletonization,hu2019topology,warping-NEURIPS2022_98143953,clCE_2024,skeleton-recall2024,lux2024topographefficientgraphbasedframework}, 
we use the two public 2D datasets, Massachusetts Roads~(Roads) \cite{dataset-MassRoads-2013} and HRF-Retina \cite{dataset-HRF-2013}. Additionally, we evaluate on two 3D datasets, the public Vessap \cite{dataset-Vessap-2019} dataset, and our CLD~(Sec.~\ref{ssec:Cement line dataset}). More details on the training splits 
are included in the supplement (Sec.~\ref{ssec: Appendix, Datasets}).
\\
\\
\textbf{Compared Methods.} 
We use the nnU-Net \cite{nnU-Isensee-2021, Ise_nnUNet_MICCAI2024}, which provides an optimized standard U-Net \cite{UnetRonneberger2015} for all experiments.
We compare CLoss against Dice \& Cross-Entropy loss (Dice \& CE), clDice \cite{clDice-Shit-2021}, and compound clDice. 
We introduce compound clDice, which corresponds to clDice but has the pixel-wise loss ($L_{pixel}$) from CLoss. 
We don't compare against \cite{clCE_2024,skeleton-recall2024,Menten2023skeletonization}, which mainly differ from clDice in $L_{pixel}$, and \cite{hu2019topology}, which has already been outperformed by clDice. 
More details on the omitted methods and method optimizations are included in the supplement~(Sec.~\ref{ssec: Appendix, Baselines}).
\\
\\
\textbf{Evaluation Metrics.} We evaluate all methods with the pixel-wise metric Dice score and the topology-aware metrics clDice \cite{clDice-Shit-2021}, Betti Number errors~(e, e$_0$, and e$_1$), e$_0$-Gt~(ours) and AGS~(ours). More details 
on the evaluation are included in the supplement~(Sec.~\ref{ssec: Appendix, Evaluation Metrics}). 
\\
\\
\textbf{Implementation Details.} 
The same skeletonization method \cite{clDice-Shit-2021}~(50 iterations) is used for all corresponding methods. 
We apply our proposed topological post-processing (Sec.~\ref{ssec:CLoss}) to all fine-tuning results to highlight the impact of our proposed loss function. Ablations for our proposed topological post-processing are provided in the supplement (Sec.~\ref{ssec: Appendix, additional quantitative results}).
We perform five-fold cross-validations for all methods. Further details are included in the supplement (Sec.~\ref{ssec: Appendix, Implementation Details}).
\subsection{Results and Discussion}
\label{ssec:Results and Discussion}
The quantitative results of our experiments are displayed in Tab.~\ref{tab:Quantitative results}. Our proposed CLoss achieves the best overall topology performance on all datasets, indicated by the combined betti error e. This is supported by the superior gap closing of CLoss for all datasets, indicated by e$_0$-Gt and AGS.
For HRF-Retina, CLoss repairs 44\,\% more missed connections than clDice~(e$_0$-Gt).
CLoss achieves the best performance for the clDice metric in all datasets except for Vessap, which can be attributed to skeletonization artifacts connected to the complex surface structure of Vessap in combination with thickened predictions from CLoss, see Fig.~\ref{fig:Metric_artifact}.
CLoss consistently outperforms compound clDice on the topology metrics, which only differs from CLoss in the critical pixel mask, and to clDice only in $L_{pixel}$. Hence, we conclude a superior critical pixel mask of our CLoss compared to clDice in the scope of our experiments.
Losses with $L_{Dice}$ for $L_{pixel}$ have a slight advantage on CLD, so we added CLoss~(Dice) for comparison with clDice, which has the same $L_{pixel}$ as clDice. 
We observe for all datasets, that superior gap closing~(better e$_0$-Gt and AGS) is linked with a stronger weighting of CLoss, which is intuitively expected.
Better overall topological performance~(e) and gap closing~(e$_0$-Gt and AGS) is also linked with a decreased Dice score for all datasets.
\begin{figure}[ht]
\centering
\includegraphics[width=0.48\textwidth]{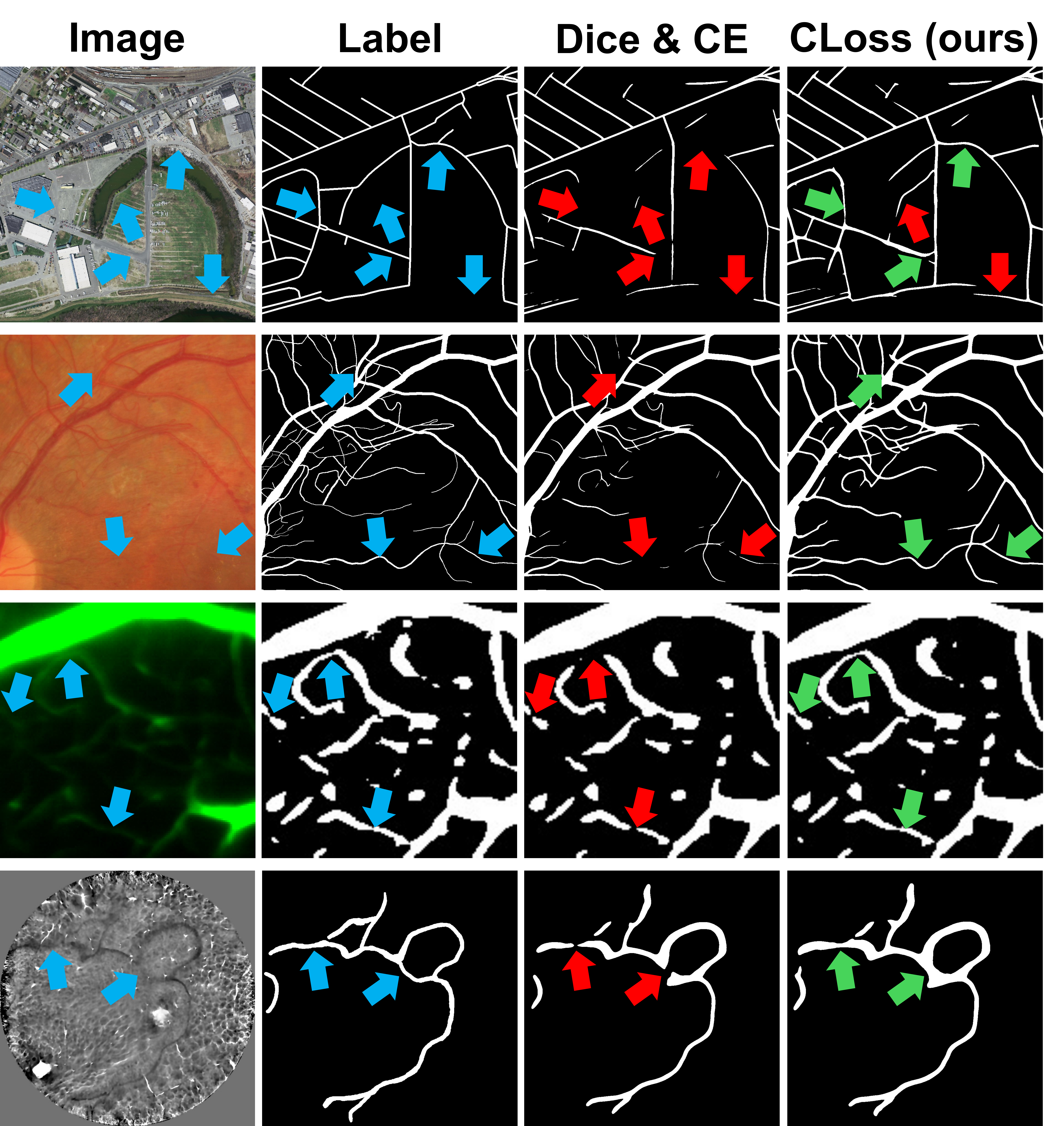}  
\caption{\textbf{Qualitative results.} The sampled patches are from Roads, HRF-Retina, Vessap, and CLD (up to down). Arrows indicate high~(green) and low~(red) topological correctness with respect to the input image/label (blue).}
\label{fig:Qualitative}
\end{figure}
\\
\\
Our quantitative results are supported by our qualitative results in Fig.~\ref{fig:Qualitative}, where we compare CLoss~($\gamma=0.2$) with Dice \& CE.
The predictions with CLoss have significantly higher topological correctness than those of Dice \& CE, highlighted by the green arrows.
For HRF-Retina, CLoss repairs especially many missed connections and doesn't introduce visible false positive connections.
All predictions of CLoss have a slightly higher thickness than the ground truth and Dice \& CE, see Fig.~\ref{fig:Qualitative}. This observed higher thickness confirms that the decreasing Dice score for higher weights of CLoss in Tab.~\ref{tab:Quantitative results} originates mainly from an increased thickness of the predictions and is compensated by increasing topological correctness.
CLoss runtime with soft skeletonization \cite{clDice-Shit-2021} is comparable to clDice, which is orders of magnitude faster than \cite{hu2019topology, warping-NEURIPS2022_98143953, lux2024topographefficientgraphbasedframework}, as compared in \cite{lux2024topographefficientgraphbasedframework}.
We include a more detailed discussion, runtimes and ablation results in the supplement~(Sec.~\ref{ssec: Appendix, Additional quantitative discussion}, Sec. \ref{ssec: Appendix, additional quantitative results}).
\section{Conclusion}
\label{sec:Conclusion}
In this work, we propose the novel topology-preserving loss function CLoss. CLoss is based on a critical pixel mask, which considers the whole context of topological errors.
We implement CLoss with a dedicated training strategy that allows for topological post-processing.
CLoss can be used on 2D and 3D datasets with any arbitrary segmentation network. Further, we benchmark on our own 3D dataset CLD, which features low contrast, diffuse borders, and image artifacts.
Additionally, we propose two intuitive metrics to verify improved connectivity due to a closing of gaps.
We demonstrate the superiority of our context-based critical pixel mask over the critical pixel mask of the entire skeletons 
on all of our four experiment datasets, where CLoss repairs up to 44\,\% more missed connections than other state-of-the-art methods.
\\
\\
\textbf{Acknowledgement.} We acknowledge the support by DASHH (Data Science in Hamburg - HELMHOLTZ Graduate
School for the Structure of Matter) with the Grant-No. HIDSS-0002.

\bibliographystyle{IEEEbib}
\bibliography{refs}

\newpage
\appendix
\section{Appendix}
\subsection{Critical pixel mask}
\label{ssec: Appendix, Critical pixel mask}
Alg.~\hyperlink{1}{1} displays the pseudo-code for the extraction of our critical pixel mask.
\begin{figure}[ht]
\captionsetup{labelformat=empty}
    \centering
    \resizebox{0.5\textwidth}{!}{
        \begin{tabular}{l}
            \hline
            \textbf{Algorithm \hypertarget{1}{1}} CLoss critical pixel mask \textbf{\textit{M}} \\
            \hline
             \textbf{Input 1:} $X$\hspace{0.8cm} \% per-pixel logits of the prediction\\
             \textbf{Input 2:} $Y$\hspace{0.8cm} \% ground truth mask\\
              $X_{bin} \leftarrow argmax(X)$ \hspace{0.8cm} \% Binarize to foreground\\
             $X_{fore} \leftarrow X($\textit{Foreground}$)$ \hspace{0.8cm} \%  Foreground likelyhood\\
             $S_{X} \leftarrow skeletonize(X_{fore})$ \hspace{0.8cm} \% Soft skeleton\\
             $S_{Y} \leftarrow skeletonize(Y)$ \\
             \\
             \textit{Determine gaps}\hspace{0.8cm} \% with distance transform (dTr)\\
             $S_{gap}\leftarrow \max(0, S_{Y} - X_{bin})$\\
             $ V_{gap}\leftarrow \bigg($dTr$\Big(1 - (S_{Y} - S_{gap})\Big)>  
             $ dTr$\Big(1 - S_{gap}\Big)\bigg)\odot Y$ \\ 
             \hspace{4cm} \% with Hadamard product $\odot$\\
             \\
             \textit{Determine false positive connections}\\
             $S_{fp} \leftarrow \max(0, S_{X} - Y)$ \\ 
             $ V_{fp}\leftarrow \bigg($ dTr$\Big(1 - (S_{X} - S_{fp})\Big)> $ dTr$\Big(1 - S_{fp}\Big) \bigg)\odot X_{bin}$ \\
             \\
             $\textbf{Output: \textit{M}} \leftarrow V_{gap}\cup V_{fp}$ \\
             \hline
        \end{tabular}
    }
\end{figure}
\\
\\
Applying soft skeletonization increases performance for binary segmentation over skeletonization of binary predictions \cite{clDice-Shit-2021, Menten2023skeletonization, skeleton-recall2024}. Our proposed method can be implemented with any improved skeleton version of \cite{clDice-Shit-2021}, eg. \cite{Menten2023skeletonization}. 
Skeletonization makes the extraction of the critical pixel mask orders of magnitudes faster than the recently proposed critical pixel mask extraction based on persistent homology \cite{hu2019topology}, as compared in \cite{lux2024topographefficientgraphbasedframework}.

\subsection{Metric Sensitivity to Gaps}
\label{ssec: Appendix Qualitative example of proposed Metrics}
Empirically, missed connections (gaps) in the predictions are 
far
more frequent than false positive connections in our used datasets, see Fig.~\ref{fig:Qualitative}.
Our proposed metrics help practitioners to identify methods that achieve superior gap closing, see Fig.~\ref{fig: Appendix Qualitative_Metric}. 
For completeness, we also emphasize the robustness of e$_0$-Gt to small separate false-positive predictions, which are a common issue for the e$_0$ metric \cite{berger2024pitfallstopologyawareimagesegmentation}.

\begin{figure}[ht] 
\centering 
\includegraphics[width=0.48\textwidth]{imgs/Qualitative_metrics.pdf}   
\caption{\textbf{Qualitative example of our proposed metrics.} Metric values are given for the displayed images. The metric values are colored for low (red\&yellow) and high (green) sensitivity for the result with better gap closing (CLoss). Our proposed metrics e$_0$-Gt and AGS help to identify the subjectively preferred CLoss result. This example is taken from the Roads dataset images in Fig.~\ref{fig:Qualitative}.}
\label{fig: Appendix Qualitative_Metric} 
\end{figure}

\subsection{Metric Susceptibility to Artifacts}
\label{ssec:micro-noise}
It is of critical importance to use evaluation metrics with a high sensitivity towards the desired topological solution \cite{Maier-Hein2024, berger2024pitfallstopologyawareimagesegmentation}.
However, popular topology metrics based on the number of connected components ($\beta_0$) are overly sensitive to artifacts, especially in the form of mico-noise~(separate components consisting of only a few pixels), which diminishes result integrity. The micro-noise problem is so far only addressed by \cite{skeleton-recall2024}, which removes small separate structures during training. Most recent solutions \cite{clDice-Shit-2021,Menten2023skeletonization, hu2019topology,warping-NEURIPS2022_98143953, clCE_2024,Li2023,lux2024topographefficientgraphbasedframework} don't address micro noise and are thereby effectively optimized for topological correctness and micro-noise suppression simultaneously.
This is a problem because we are primarily interested in developing methods to achieve major topological correctness and not to achieve micro-noise suppression. We argue that micro-noise suppression should only be seen as a desired property of a method but not as a primary attribute to compare different methods for major topological correctness. We propose the following solution.
\\
\\
\textbf{Proposed Topological Post-Processing.}
We propose to leverage the natural characteristic of overlap-based losses to minimize artifacts (and micro-noise) \cite{Do-we-really-need-dice2024,MA2021102035}.
Therefore, we propose to include a pretraining with a topology-insensitive overlap-based loss function by default and then conduct a fine-tuning with a topology-aware loss function. 
The suggested pretraining is compatible with all existing topology-preserving loss functions. 
We suggest to only keep the predictions of the fine-tuning training ($X_{bin}^{ft}$), which are connected to the pretraining predictions ($X_{bin}^{pre}$).
Expressed as an equation, this means for the mask $c_i$ of each connected component $i$ in $X_{bin}^{ft}$ we suggest to only keep $i$ in $X_{bin}^{ft}$ if 
\begin{equation}
sum \big(c_i \odot X_{bin}^{pre} \big) > 0 
\label{eq: micro-noise}
\end{equation}
In particular, our proposed topological post-processing doesn't alter the connectivity of $X_{bin}^{opt}$ in any way except for removing separate structures which were not present in the pretraining. Our suggested post-processing supports the intuition of improving the topology of an existing structure without adding new structures.
\\
\\
We note that a comparable effect of micro noise removal can in theory be achieved with a post-processing of opening and erosion operations. However, the applicability of opening and erosion operations depends highly on the predicted structures. Opening and erosion operations have a high risk of introducing new false positive connections that weren't present before the post-processing. Our suggested post-processing is by design robust to introducing new artifacts in the form of false positive connections.
\\
\\
Compared to the runtime solution of \cite{skeleton-recall2024}, our approach doesn't require any additional hyperparameters to select the size of the removed structures and will also remove artifacts which are missed by a fixed threshold.

\subsection{Cement line dataset (CLD)}
\label{ssec: Appendix, Cement line dataset}
\begin{figure}[ht] 
\centering   
\includegraphics[width=0.48\textwidth]{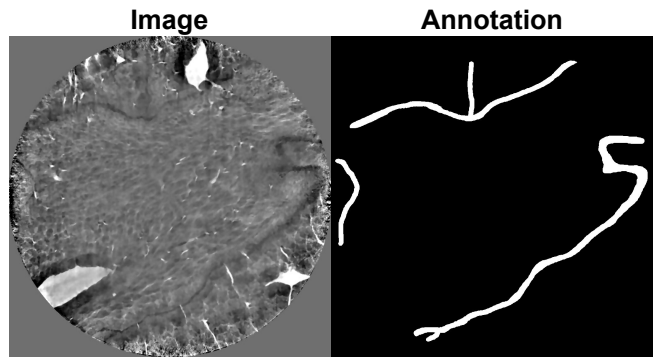}
\caption{\textbf{Cement line Annotation.} The cement lines have low contrast and diffuse borders. Also, their appearance is diffuse, note in the left image the changing grayscale characteristics of the cement line in the lower right corner
from bottom to right.}
\label{fig:CLD}
\end{figure}

\textbf{Image Acquisition.} Specific steps for acquiring the used 3D images of the nanotomography bone cement line dataset are described in \cite{CL-SEFA2023154}, sections 2.4.2 to 2.4.4.1. These steps include the preparation of the imaged specimen, the imaging procedure and the processing of the raw images. We cropped the images from their initial shape 1024² $\times$ 1024 to their final shape 1024² $\times$ 600 to focus on areas with pronounced cement lines and minimal artifacts. 
\\
\\
\textbf{Annotation.} The annotation was done for each image on the complete initial shape of 1024² $\times$ 1024. The annotation of the cement lines was done manually in Avizo~(version 2024.2, Thermo Fisher Scientific, Waltham, USA), utilizing the standard brush and interpolation tools. The cement lines have low contrast and diffuse borders. Importantly, their visibility and 
area proportion in the 2D view of the 3D image critically depends on the view angle in 3D, which doesn't always align well with the available view axes (xy,xz,zy). When the cement lines slightly change orientation inside the bone they can become much less pronounced in the current 2D view direction and significantly change in the displayed size. 
\\
\\
To address these issues, we conducted a pre-annotation for all 3D images from all three view angles (axes:xy,xz,zy) to acquire a coarse initial representation of all cement lines. The pre-annotation corresponds to a detailed annotation about every 50 slices for each view axis for every 3D image. The pre-annotation was used to determine the preferred orientation of the majority of the cement lines within the 3D image. The pre-annotation was further used as an aid in the more thorough annotation to not miss any cement lines due to viewing angle effects. The thorough annotation was done along the most suitable view axis (xy,xz, or zy). This most suitable view axis was chosen in a way that the majority of the cement lines had a small area proportion in the 2D view of the 3D image, see Fig.~\ref{fig:CLD}. The thorough annotation corresponds to a detailed annotation about every three to five slices along the most suitable view axis for every 3D image. The thorough annotation for each 3D image was then interpolated with the standard Avizo interpolation tool. This interpolation was designed by Avizo to only interpolate between annotation slices without considering any gray values. However, the resulting annotation was a suitable compromise between annotation time and annotation accuracy. 
\\
\\
Sometimes, the interpolation tool didn't work as expected, and there were holes within the annotation that didn't get interpolated. To close these interpolation holes, we conducted a series of morphological opening and closing operations until the holes were closed. We repeated this step with additional annotated slices when this approach didn't work initially. We note that the interpolation had also severe trouble interpolating more than one cement line at a time. As a solution, we conducted the thorough annotation and interpolation procedure only for selected parts of a single cement line at once, which we later combined into the complete annotation.

\subsection{Datasets}
\label{ssec: Appendix, Datasets}
Roads contains aerial images~(1500²) of the road system in Massachusetts, similar to settings in \cite{clDice-Shit-2021} and \cite{warping-NEURIPS2022_98143953} we use 421 images for training and 49 for testing, after excluding images with a white-masked pixel percentage of more than 4\%. HRF-Retina consists of high-resolution fundus images~(3504 $\times$ 2336) for retinal vessel segmentation, we use 36 images for training and 9 for testing. Vessap contains volumetric scans~(500² $\times$ 50) of brain vessels in two channels, similar to settings in \cite{clDice-Shit-2021} we use 8 3D images for training and 3 for testing, using both input channels simultaneously. Our CLD contains volumetric scans~(1024² $\times$ 600) of bone cement lines, we use 13 3D images for training and 4 for testing. We conduct a five-fold cross-validation for all datasets. 
\\
\\
We further note that the high-resolution of HRF-Retina is especially suited for examining the influence of the critical pixel mask as the topological errors are represented by more pixels than in lower-resolution images. 

\subsection{Compared Methods}
\label{ssec: Appendix, Baselines}
The nnU-Net \cite{nnU-Isensee-2021} is a framework that provides an improved version of the standard U-Net \cite{UnetRonneberger2015} with optimized hyperparameters. The nnU-Net is by default trained with a compound loss, consisting of Dice and Cross-Entropy loss \cite{MA2021102035}. \\
\\
clDice \cite{clDice-Shit-2021} is another topology-aware segmentation loss based on critical pixels that focuses on the extracted skeleton of likelihood maps and ground truth masks. \\
\\
\textbf{Method optimization.} To demonstrate the effectiveness and better topological performances of our proposed method, we ensure identical prerequisites of our proposed method and the compared methods so that better performance can be clearly attributed to our proposed critical pixel mask. In that spirit, we conduct hyperparameter optimizations not only for our proposed method but also for the compared methods. 
For compound clDice, we conducted a weight hyperparameter search for each dataset ranging from 0.1 to 0.5 in 0.1 steps. For space reasons, we only report the compound clDice result with the highest clDice metric in Tab.~\ref{tab:Quantitative results}. For clDice, we use the weight with the highest clDice metric for each dataset from their paper. For CLD, we choose the same weight for clDice as for the optimized compound clDice.
\\
\\
\textbf{Omitted Methods.} We compare against clDice loss \cite{clDice-Shit-2021}, a state-of-the-art topology-aware 
segmentation loss, which defines the skeleton of predictions and ground truth as critical pixel mask. clDice and \cite{clCE_2024} only differ in $L_{pixel}$. clDice and \cite{Menten2023skeletonization} only differ in the skeleton algorithm, having the same  $L_{pixel}$ and critical pixel mask. \cite{skeleton-recall2024} is the multi-class adaption of clDice with a different $L_{pixel}$ and a reduced critical pixel mask in comparison to clDice that only includes the slightly dilated ground truth skeleton for computational efficiency. Additionally, \cite{skeleton-recall2024} removes small structures during training which makes it difficult to point out from which design-choice their reported improvement over clDice (no removed structures) originates. However, as \cite{skeleton-recall2024} note themselves an inferior skeleton extraction for binary segmentation compared to clDice, we suspect their critical pixel mask selection to be inferior to clDice in binary segmentation as it doesn't consider false positive connections.
\\
\\
Our proposed method can be adapted with any skeleton extraction and $L_{pixel}$. Hence, we focus on validation against methods with a distinctly different critical pixel mask selection strategy. Therefore, we don't validate against \cite{Menten2023skeletonization, clCE_2024} and \cite{skeleton-recall2024}, which use different skeletonization methods or $L_{pixel}$ than clDice but not superior critical pixel masks. CLoss outperforming compound clDice implies CLoss outperforming \cite{Menten2023skeletonization, clCE_2024, skeleton-recall2024}.
\\
\\
We would have liked to compare against \cite{warping-NEURIPS2022_98143953} but weren't able to reproduce their results.
We could only reproduce their proposed critical pixel mask for the 2D datasets but not on our 3D CLD. We suspect that this might be caused by the complex 3D surface structure in our CLD. Initial training attempts with the original code from their repository resulted in numerous errors for the 2D and 3D datasets, which we could only fix for the 3D datasets. However, related work \cite{lux2024topographefficientgraphbasedframework, Li2023, skeleton-recall2024, clCE_2024} also doesn't report results on \cite{warping-NEURIPS2022_98143953}, so the implementation of \cite{warping-NEURIPS2022_98143953} might be a common issue.
\\
\\
There also exist numerous other topology-preserving approaches that we don't compare against, since this would go well beyond the scope of this paper. These other approaches are not based on critical pixel masks and include for example graph-based 
approaches \cite{lux2024topographefficientgraphbasedframework} (\cite{lux2024topographefficientgraphbasedframework} only works with 2D data) or approaches entirely based on post-processing \cite{Li2023} (omits to improve topology correctness already in the image domain). We note that \cite{Li2023} can be complemented with our proposed method.

\subsection{Evaluation Metrics}
\label{ssec: Appendix, Evaluation Metrics}
\textbf{Evaluation patch sizes.} We evaluated the Dice metric respectively over the whole test samples for the used 2D and 3D datasets.\\
\\
As mentioned in Sec.~\ref{ssec:Proposed Metrics}, e$_0$ is not able to capture gaps if it is computed over the whole image for datasets with high overall connectivity. 
We calculate the metrics on patch sizes which provide meaningful evaluation for the used datasets. We iterate over the whole image shape with the patches as sliding window (similar to \cite{clCE_2024}) instead of sampling patches randomly for evaluation like \cite{hu2019topology,warping-NEURIPS2022_98143953, clDice-Shit-2021, lux2024topographefficientgraphbasedframework}.
\\
\\
For 2D datasets, we evaluate clDice and AGS on the whole image. We calculate betti-metrics and e$_0$-Gt on patches, 375²~(Roads) and 292²~(HRF-Retina).\\
\\
For the 3D dataset Vessap, we calculate clDice, AGS and e$_0$-Gt over the whole sample volume. We calculate betti-metrics on full image size along the z-direction.\\
\\
For our 3D CLD, we calculate clDice, AGS and e$_0$-Gt over patches of 1024² $\times$ 64, where we also consider corrections of the clDice metric for its shortcoming with empty patches. We note that this patching in comparison to Vessap is due to memory reasons. Further, we evaluate betti-metrics on full image size along the z-direction.

\subsection{Implementation Details}
\label{ssec: Appendix, Implementation Details}
We conduct all our trainings within the nnU-Net framework \cite{nnU-Isensee-2021} to ensure a maximum of reproducibility and leading performance \cite{Ise_nnUNet_MICCAI2024}.
The nnU-Net standard training has a length of 1000 epochs and utilizes Dice and Cross-Entropy as equally weighted compound loss. Training is done with a five-fold cross-validation for all methods. Inference is by standard done with all five folds simultaneously as an ensemble.\\
\\
For our 2D datasets, we use the nnU-Net configuration 2d. For our 3D datasets, we use the configuration 3d\_fullres.
\\
\\
We used the standard nnU-Net configuration for all methods, meaning our results can be reproduced with any GPU that has more than 11\,GB VRAM. For increased speed, we conduct all trainings on A100 GPUs. As our used datasets are comparably small, we use the standard nnU-Net configuration without the residual encoder presets. We use PyTorch framework version 2.3.0 to implement our proposed method.
\\
\\
The nnU-Net configured a batch size of two for all datasets. The patch sizes for the 2D datasets were configured by nnU-Net to 1280 $\times$ 1024~(Roads), 1536 $\times$ 1024~(HRF-Retina) and for the 3D datasets to 256² $\times$ 32~(Vessap) and 160² $\times$ 90~(CLD).
\\
\\
For connectivity, we apply 8-adjacency in 2D and 26-adjacency in 3D.
\\
\\
\textbf{Inference.} 
Alg.~\hyperlink{2}{2} displays the pseudo-code for the inference with our proposed training-regime from Fig.~\ref{fig:CLoss}.
\begin{figure}[ht]
\captionsetup{labelformat=empty}
    \centering
    \resizebox{0.5\textwidth}{!}{
        \begin{tabular}{l}
            \hline
            \textbf{Algorithm \hypertarget{2}{2}} CLoss inference\\
            \hline
             \textbf{Input:} $D$ \hspace{4cm} \% domain image\\
              $X_{bin}^{pre} \leftarrow predictions_{Pretraining}(D)$ \hspace{0.1cm} \% Binarized\\
              \\
              $X_{bin}^{ft} \leftarrow predictions_{fine-tuning}(D)$ \hspace{0.1cm} \% Binarized\\
            \\
             $X_{bin} \leftarrow post$-$processing(X_{bin}^{pre}, X_{bin}^{ft})$ 
             \\
             \hspace{4cm} \% Only keeps structures of $X_{bin}^{ft}$ \\
             \hspace{4cm} already present in $X_{bin}^{pre}$
             \\
            \\
             $\textbf{Output: \textit{$X_{bin}$}}$ \hspace{2cm} \% Binarized final predictions\\
             \hline
        \end{tabular}
    }
\end{figure}
\\
\\
\textbf{Guideline for tuning gamma.} If $\gamma$ tends to 0, CLoss will converge to Dice\&CE loss. A higher $\gamma$ encourages more topological correctness. It should be $\gamma<0.5$ to not focus overly on topological errors. All our datasets show good results for a value of 0.2. For new datasets, we recommend the common exploration approach in multiples of 3, e.g. 0.3, 0.1, 0.03.
\\
\\
\textbf{Comment on generalization of CLoss to object segmentation of tumours and organs.}
CLoss promotes topology-preserving segmentation along the skeleton-pixels of the predictions and ground truth. Therefore, CLoss can be beneficial in all areas where the skeleton-pixels yield desirable information. We note that skeleton-pixels are typically not important for spherical objects like the human heart but are of increased importance for elongated objects like the trachea. 

\subsection{Additional Discussion of Quantitative Results }
\label{ssec: Appendix, Additional quantitative discussion}
\textbf{HRF-Retina.} CLoss has the smallest error on the number of connected components (e$_0$) in all datasets except for HRF-Retina.
We attribute the value for HRF-Retina to the mentioned susceptibility of e$_0$ to artifacts (Sec.~\ref{ssec:Proposed Metrics}) from the patch-based evaluation.
The artifact distortion of e$_0$ is indicated by e$_0$-Gt, AGS, and the clDice metric. Our proposed metrics e$_0$-Gt and AGS clearly show a better gap closing of CLoss. Additionally, also the clDice metric of CLoss for HRF-Retina is significantly improved (about 2\%) over its closest competing method. The clDice metric is especially reliable for HRF-Retina due to the smooth surface structure of the vessels. This indicates a better topology performance of CLoss for HRF-Retina despite the seemingly unfavorable e$_0$ value. 
\\
\\
\textbf{Vessap.} For the Vessap dataset, we observe that the best clDice score is achieved by the topology-insensitive nnU-Net~(pretraining). This seems to contradict with the other topology-sensitive metrics. Therefore, we suspect that the thin network 3D structures of Vessap could lead to many seemingly false positive predictions of the prediction skeleton, which distorts the topological sensitivity of the clDice metric on this dataset, see Fig.~\ref{fig:Metric_artifact}. This hypothesis is supported by an increasing AGS score for the topological fine-tuned methods. We also note that the seemingly high e$_0$ values for Vessap come from our patch-based evaluation of~(full 2D image sizes stepping along the z-direction) in combination with the 3D network structure of Vessap. We note that e$_1$ in this context can especially contain artifacts from the slicing. Therefore, a minimized e$_0$ indicates topological performance on Vessap more reliable than minimized e$_1$.
\\
\\
\textbf{CLD.} The unmodified clDice loss only has a better gap closing~(better e$_0$-Gt and AGS) than our standard CLoss on CLD. Importantly, this is only due to a different pixel-wise loss and not due to the different critical pixel mask. We verify this by adding results of CLoss~(Dice) to CLD, which has the same pixel-wise loss function as clDice and only differs in the critical pixel mask to the clDice loss. CLoss~(Dice) visibly outperforms clDice on the topology metrics. Interestingly, CLoss~(Dice) is better than our proposed standard CLoss in e$_1$, e$_0$-Gt, and AGS but not in clDice, e, and e$_0$. This indicates that the target structure of our CLD favors different $L_{pixel}$ for different metrics, which is not the case for the other datasets (compound clDice consistently better or comparable to original clDice). The favoring of different $L_{pixel}$ for different metrics further illustrates the increased complexity of CLD over the other datasets.

\subsection{Ablations}
\label{ssec: Appendix, additional quantitative results}
\textbf{Post-Processing.} Additional results of the post-processing are displayed in Tab.~\ref{tab:Appendix_pp_ablations}.
All metric values in this example improve except for e$_1$, which remains the same, and AGS, which slightly decreases. All methods benefit from our post-processing in a similar quantity. It can be observed for all results, that the change in the Dice value is comparably small to the change in $e$. This indicates, that our post-processing primarily removes noise in the form of small separated structures.
\begin{table*}[ht]
    \caption{\textbf{Post-Processing.} Example for CLD.}
    \label{tab:Appendix_pp_ablations}
    \centering
    \begin{tabular}{ccc|ccccccc}
        \hline
        \hline
        Post-Processing&Method & Weight $\gamma$ & Dice↑ & clDice↑ \cite{clDice-Shit-2021} & e↓ & e$_1$↓ & e$_0$↓ & \textit{e$_0$-Gt}↓ & \textit{AGS}↑ \\
        \hline
        \hline
         w/o&Dice \& CE &  & 70.23 & 85.17 & 3.615 & 1.075 & 2.540 & 2.479 & 82.03\\
         with& & & 70.27 & 85.23 & 3.422 & 1.075 & 2.347 & 2.447 & 82.02\\
         \hline
         w/o&Compound clDice & 0.5 & 70.58 & 85.39 & 3.325 & 1.065 & 2.259 & 2.176 & 82.67\\
         with& &  & \textbf{70.61} & 85.45 & 3.165 & 1.065 & 2.099 & 2.150 & 82.66\\
         \hline
         w/o&\textit{CLoss} & \textit{0.2} & 70.41 & 86.75 & 3.224 & \textbf{1.047} & 2.177 & 1.916 & \textbf{85.99}\\
         with&& & 70.44 & \textbf{86.83} & \textbf{3.020} & \textbf{1.047} & \textbf{1.973} & \textbf{1.899} & 85.98\\
        \hline
        \hline
    \end{tabular}
\end{table*}
\\
\\
\textbf{Fine-tuning length.} Ablations for the fine-tuning length are displayed in Tab.~\ref{tab:Appendix_Epochs_ablations}. A shorter fine-tuning length seems preferable, although we note that there is no consistent trend between the different lengths. For more detailed ablations on the fine-tuning between 50 epochs and 100 epochs, there was no clear winner, as might be falsely suggested by Tab.~\ref{tab:Appendix_Epochs_ablations}. Hence, we chose 50 epochs for computational efficiency.
\begin{table*}[ht]
    \caption{\textbf{Fine-tuning length.} Example for CLD. All results are post-processed.}
    \label{tab:Appendix_Epochs_ablations}
    \centering
    \begin{tabular}{ccc|ccccccc}
        \hline
        \hline
        Epochs&Method & Weight $\gamma$ & Dice↑ & clDice↑ \cite{clDice-Shit-2021} & e↓ & e$_1$↓ & e$_0$↓ & \textit{e$_0$-Gt}↓ & \textit{AGS}↑ \\
        \hline
        \hline
         50&Compound clDice & 0.5 & 70.61 & 85.45 & 3.165 & 1.065 & 2.099 & 2.150 & 82.66\\
         100& & & 70.67 & 85.37 & 3.108 & 1.062 & 2.046 & 2.086 & 82.79\\
         150& &  & 70.58 & 85.22 & 3.097 & 1.064 & 2.034 & 2.180 & 82.19\\
         300& &  & 70.65 & 85.44 & 3.115 & 1.053 & 2.062 & 2.097 & 82.71\\
         \hline
         50&\textit{CLoss}&\textit{0.1}  & 70.75 & \textbf{86.26} & 3.109 & 1.069 & 2.040 & 2.040 & 84.44\\
         100&&  & \textbf{70.77} & 86.25 & \textbf{3.046} & \textbf{1.050} & \textbf{1.996} & \textbf{2.010} & \textbf{84.78}\\   
         150&&  & 70.46 & 85.64 & 3.173 & 1.062 & 2.110 & 2.218 & 83.02\\
         300&&  & 70.51 & 85.34 & 3.075 & 1.060 & 2.015 & 2.079 & 82.39\\
        \hline
        \hline
    \end{tabular}
\end{table*}
\\
\\
\textbf{Critical pixel mask.} Additional ablations for the critical pixel mask are included in Tab.~\ref{tab:Appendix_critical_pxl_mask_ablations}. We include Thin CLoss, which has a critical pixel extraction analog to CLoss except for the context extraction. Therefore, Thin CLoss only considers the skeleton at the topological errors. Our proposed critical pixel mask with context extraction yields significantly better topological correctness. CLoss seems only second best in $e_1$ to compound clDice ($\gamma=0.1$), but the other bad metric values of compound clDice ($\gamma=0.1$) indicate that this is likely not due to overall topological correctness but rather related to artifacts.  
\begin{table*}[ht]
    \caption{\textbf{Critical pixel mask.} Example for CLD. All results are post-processed. All methods differ only in the critical pixel mask.}
    \label{tab:Appendix_critical_pxl_mask_ablations}
    \centering
    \begin{tabular}{ccc|ccccccc}
        \hline
        \hline
        Critical pixel mask & Method & Weight $\gamma$ & Dice↑ & clDice↑ \cite{clDice-Shit-2021} & e↓ & e$_1$↓ & e$_0$↓ & \textit{e$_0$-Gt}↓ & \textit{AGS}↑ \\
        \hline
        \hline
         w/o&Dice \& CE & & 70.27 & 85.23 & 3.422 & 1.075 & 2.347 & 2.447 & 82.02\\
         \hline
         Full skeleton&Compound clDice & 0.1 & 68.93 & 83.11 & 3.522 & \textbf{1.029} & 2.493 & 2.679 & 79.06\\
         && 0.2 & 70.47 & 84.96 & 3.361 & 1.077 & 2.283 & 2.352 & 81.81\\
         && 0.3 & 70.61 & 85.41 & 3.230 & 1.077 & 2.153 & 2.249 & 82.80\\
         && 0.4 & 70.51 & 85.16 & 3.287 & 1.063 & 2.223 & 2.283 & 82.26\\
         && 0.5 & 70.61 & 85.45 & 3.165 & 1.065 & 2.099 & 2.150 & 82.66\\
         \hline
         Skeleton at topological errors&\textit{Thin CLoss} & 0.08 & 70.33 & 85.06 & 3.353 & 1.087 & 2.266 & 2.438 & 81.53\\
         && \textit{0.2} & 69.76 & 84.39 & 3.356 & 1.095 & 2.261 & 2.492 & 79.66\\
         && \textit{0.5} & 69.62 & 84.05 & 3.475 & 1.091 & 2.383 & 2.643 & 79.29\\
         \hline
         Context at topological errors&\textit{CLoss} & \textit{0.08} & \textbf{70.98} & 86.22 & 3.158 & 1.063 & 2.095 & 2.061 & 84.47\\
         && \textit{0.1} & 70.75 & 86.26 & 3.109 & 1.069 & 2.040 & 2.040 & 84.44\\
         && \textit{0.2} & 70.44 & \textbf{86.83} & \textbf{3.020} & 1.047 & \textbf{1.973} & \textbf{1.899} & \textbf{85.98}\\
         \hline
         \hline
    \end{tabular}
\end{table*}
\\
\\
\textbf{Pixel-wise loss.} An ablation on $L_{pixel}$ is included in Tab.~\ref{tab:Appendix_Lpxl_ablations}. clDice and compound clDice differ only in $L_{pixel}$. clDice has $L_{Dice}$, and compound clDice the equally weighted combination of $L_{Dice}$ and $L_{CE}$, analogue to our CLoss implementation. 
Compound clDice performs better on average for Roads, HRF-Retina, and Vessap. For CLD we note an advantage for clDice.
\begin{table*}[ht]
    \caption{\textbf{Pixel-wise loss.} We focus on ablation between $L_{Dice}$ and our $L_{pixel}$ used for CLoss. All results are post-processed.}
    \label{tab:Appendix_Lpxl_ablations}
    \centering
    \begin{tabular}{ccc|ccccccc}
        \hline
        \hline
        Dataset & Method & Weight $\gamma$ & Dice↑ & clDice↑ \cite{clDice-Shit-2021} & e↓ & e$_1$↓ & e$_0$↓ & \textit{e$_0$-Gt}↓ & \textit{AGS}↑ \\
        \hline
        \hline
         \multirow{2}{*}{\centering Roads} 
         &clDice \cite{clDice-Shit-2021} & 0.5 & 79.50 & 89.11 & 1.126 & 0.897 & 0.230 & 0.710 & 87.07\\
         &Compound clDice & 0.5 & \textbf{79.59} & \textbf{89.21} & \textbf{1.065} & \textbf{0.870} & \textbf{0.195} & \textbf{0.673} & \textbf{87.25}\\
        \hline
        \hline
         \multirow{2}{*}{\centering HRF-Retina}
         &clDice \cite{clDice-Shit-2021} & 0.5 & 82.15 & \textbf{83.09} & 0.426 & 0.256 & 0.170 & 2.475 & \textbf{82.23}\\
         &Compound clDice & 0.5 & \textbf{82.33} & 82.98 & \textbf{0.405} & \textbf{0.250} & \textbf{0.155} & \textbf{2.429} & 81.25\\
        \hline
        \hline
         \multirow{2}{*}{\centering Vessap}
         &clDice \cite{clDice-Shit-2021} & 0.4 & 92.70 & 94.86 & 29.00 & 1.240 & 27.76 & \textbf{9.44} & \textbf{97.74}\\
         &Compound clDice & 0.4 & \textbf{92.98} & \textbf{95.42} & \textbf{26.880} & \textbf{1.220} & \textbf{25.660} & 11.040 & 97.28\\
        \hline
        \hline
         \multirow{2}{*}{\centering \textit{CLD}} 
         &clDice \cite{clDice-Shit-2021} & 0.5 & \textbf{70.88} & \textbf{86.22} & 3.374 & \textbf{1.042} & 2.333 & \textbf{1.825} & \textbf{87.11}\\
         &Compound clDice & 0.5 & 70.61 & 85.45 & \textbf{3.165} & 1.065 & \textbf{2.099} & 2.150 & 82.66\\
         \hline
         \hline
    \end{tabular}
\end{table*}
\\
\\
\textbf{Compound clDice optimization.}
Additional results of the compound clDice optimization are displayed in Tab.~\ref{Tab:Appendix, Roads compclDice ablations}.
CLoss yields consistently better results.
\begin{table*}[ht]
    \caption{\textbf{Compound clDice optimization.} Example for Roads dataset. All results are post-processed.}
    \label{Tab:Appendix, Roads compclDice ablations}
    \centering
    \begin{tabular}{cc|ccccccc}
        \hline
        \hline
        Method & Weight $\gamma$ & Dice↑ & clDice↑ \cite{clDice-Shit-2021} & e↓ & e$_1$↓ & e$_0$↓ & \textit{e$_0$-Gt}↓ & \textit{AGS}↑ \\
        \hline
        \hline
         nnU-Net \cite{nnU-Isensee-2021} &  & 79.69 & 89.34 & 1.181 & 0.895 & 0.286 & 0.702 & 86.46\\
         Dice \& CE &  & 79.77 & 89.37 & 1.156 & 0.949 & 0.207 & 0.699 & 86.41\\
         \hline
         Compound clDice& 0.1 & 79.69 & 89.26 & 1.089 & 0.866 & 0.223 & 0.691 & 86.78\\
         & 0.2 & 79.59 & 89.12 & 1.082 & 0.880 & 0.202 & 0.708 & 86.87\\
         & 0.3 & 79.73 & 89.34 & 1.108 & 0.909 & 0.199 & 0.673 & 87.05\\
         & 0.4 & 79.68 & 89.26 & 1.125 & 0.915 & 0.210 & 0.677 & 87.05\\
         & 0.5 & 79.59 & 89.21 & 1.065 & 0.870 & 0.195 & 0.673 & 87.25\\
         \hline
         \textit{CLoss} & \textit{0.08} & \textbf{79.82} & \textbf{89.47} & 1.065 & 0.880 & 0.185 & 0.656 & 87.70\\
         & \textit{0.1} & 79.57 & 89.21 & \textbf{0.990} & 0.810 & \textbf{0.180} & 0.617 & 87.88\\
         & \textit{0.2} & 79.12 & 89.13 & 0.994 & \textbf{0.788} & 0.205 & \textbf{0.494} & \textbf{89.48}\\
        \hline
        \hline
    \end{tabular}
\end{table*}
\\
\\
\textbf{Runtime.} Results on the runtime are displayed in Tab.~\ref{tab:Runtime}. The runtime of CLoss is comparable to clDice. In perspective, other state-of-the-art approaches  \cite{hu2019topology, warping-NEURIPS2022_98143953, lux2024topographefficientgraphbasedframework} have 10 to 60 times longer loss calculations than clDice, as compared by \cite{lux2024topographefficientgraphbasedframework}. Generating predictions on the test set is independent of the utilized loss and takes about 3min for the Roads dataset.
\begin{table*}[ht]
    \caption{\textbf{Runtime.} Example for Roads dataset.}
    \label{tab:Runtime}
    \centering
    \begin{tabular}{ccc}
        \hline
        Loss & Time/epoch & Time/training\\
        \hline
        \hline
        Dice \& CE & \textbf{22s} & \textbf{29min}\\
        clDice & 63s & 59min\\
        \textit{CLoss} & 78s & 71min\\
        \hline
    \end{tabular}
\end{table*}

\end{document}